# 基于多模态深度融合网络的无人机轨迹预测

高塬[1]，郭馨雨[1]，谢文静[1]，王子凡[2]，余鸿文[1]，李恭杨[1]，徐树公[3]
(1. 上海大学通信与信息工程学院，上海，200444；2. 西南大学计算机与信息科学学院，重庆，400715 ; 3. 西交利物浦大学智能工程学院，苏州，215123.)

**摘　要**　为满足低空经济中对于未授权无人机（Unmanned Aerial Vehicle, UAV）的管理，提出一种激光雷达与毫米波雷达信息多模态融合的无人机轨迹预测方法方案。设计了一种用于多模态无人机轨迹预测的深度融合网络（Multi-Modal Deep Fusion Framework），整体结构由两种模态的特征提取网络与双向交叉注意力融合模块两部分组成，旨在充分利用 LiDAR 与 Radar 点云在空间几何结构与动态反射特性上的互补信息。在特征提取阶段，模型分别为 LiDAR 与 Radar 设计了独立但结构相同的特征编码器，提取特征之后，模型进入双向交叉注意力机制（Bidirectional Cross-Attention Mechanism）阶段，以实现两种模态间的信息互补与语义对齐。为验证本文提出模型的有效性，采用 CVPR2024 在 UG2+UAV Tracking and Pose-Estimation Challenge 比赛中所使用的 MMAUD 数据集作为训练集与测试集，实验结果显示本文所提出的多模态融合模型显著提升了轨迹精度，相比于基线模型精度提升 40%。并且通过消融实验，论证了不同损失函数、后处理策略对于提升模型性能的有效性。此模型能够有效利用多模态数据，为低空经济中非授权无人机轨迹预测提供一种高效的解决方案。

# Efficient UAV Trajectory prediction: A multi-modal deep diffusion framework

GAO Yuan[1], GUO Xinyu[1], XIE Wenjing[1]，WANG Zifan[2], YU Hongwen[1], LI Gongyang[1], XU Shugong[3]
(1. School of Communication and Information Engineering, Shanghai University, Shanghai, 200444；2. College of Computer and Information Science, Southwest University, Chongqing, 400715; 3. Department of Intelligent Science, Xi'an Jiaotong-Liverpool University, Suzhou, 215123.)

**Abstract**　To meet the requirements for managing unauthorized UAVs in the low-altitude economy, a multi-modal UAV trajectory prediction method based on the fusion of LiDAR and millimeter-wave radar information is proposed. A deep fusion network for multi-modal UAV trajectory prediction, termed the Multi-Modal Deep Fusion Framework, is designed. The overall architecture consists of two modality-specific feature extraction networks and a bidirectional cross-attention fusion module, aiming to fully exploit the complementary information of LiDAR and radar point clouds in spatial geometric structure and dynamic reflection characteristics. In the feature extraction stage, the model employs independent but structurally identical feature encoders for LiDAR and radar. After feature extraction, the model enters the Bidirectional Cross-Attention Mechanism stage to achieve information complementarity and semantic alignment between the two modalities. To verify the effectiveness of the proposed model, the MMAUD dataset used in the CVPR 2024 UG2+ UAV Tracking and Pose-Estimation Challenge is adopted as the training and testing dataset. Experimental results show that the proposed multi-modal fusion model significantly improves trajectory prediction accuracy, achieving a 40% improvement compared to the baseline model. In addition, ablation experiments are conducted to demonstrate the effectiveness of different loss functions and post-processing strategies in improving model performance. The proposed model can effectively utilize multi-modal data and provides an efficient solution for unauthorized UAV trajectory prediction in the low-altitude economy.

资助项目：国家自然科学基金(62501373); 国家自然科学基金(62401350)；上海市自然科学基金项目(25ZR1402148) ；上海市科委启明星项目扬帆专项(24YF2713000)；
Foundation Items：National Natural Science Foundation of China(62501373); National Natural Science Foundation of China(62401350);　Shanghai Natural Science Foundation (25ZR1402148)；Shanghai Sailing Program (24YF2713000);
*通信作者：高塬, E-mail: gaoyuansie@shu.edu.cn. Corresponding Author: GAO Yuan, E-mail: gaoyuansie@shu.edu.cn.

## 0 引 言

随着无人机（Unmanned Aerial Vehicle, UAV）技术的迅猛发展，无人机作为一种通过无线电遥控或自主程序飞行的空中载具，具有部署便捷、机动性强以及成本低廉等显著优势，在军事、民用和商业领域展现出巨大价值[1]-[5]。在军事领域，无人机可执行侦察、目标定位、情报搜集以及诱饵火力等任务[6],[7]；在民用领域，可完成快递运输、农药喷洒、天线巡检等高空和高风险作业；在商业领域，则可实现多角度航拍以捕捉更优视角的影像[8]。然而，随着无人机领域的快速崛起，其安全问题日益凸显，未经授权的无人机活动日益增多，对机场、军事基地、核电站等关键基础设施以及公共隐私构成了严重威胁[9],[10]。例如，非授权无人机可能携带爆炸物、进行间谍活动、干扰正常航空秩序，甚至用于走私或对监狱进行监视，从而导致航班延误、安全事故或国家安全风险。近年来，全球范围内多次发生无人机入侵事件，如机场因入侵而关闭，造成数百万美元的经济损失，或有无人机在边境地区用于监视和入侵活动。这些事件凸显了低空空域安全管理的紧迫性。因此，未授权无人机的轨迹识别已成为该领域关键研究方向。通过实时检测、跟踪和预测无人机的飞行路径，可以有效识别潜在风险、实施反制措施，并保障空域安全。在此过程中，轨迹预测尤为重要，它不仅能预判非授权无人机的潜在威胁路径，还能为安全监督系统提供决策依据，实现从被动响应向主动预防的转变[11]。

目前，在未授权无人机轨迹预测方面，主要研究内容包含无人机识别、类型分类、轨迹预测等。对于上述任务，单模态往往无法完美的完成任务[17]。基于视觉的无人机定位方法在目标识别和类别区分方面具有一定优势[15]，但其性能高度依赖外界光照条件和成像质量。当无人机处于远距离、小视角或被建筑物、植被等遮挡时，视觉传感器往往难以获得连续、稳定的观测信息。此外，在夜间、逆光、雾霾或雨雪等极端环境下，视觉特征显著退化，目标易出现丢失或误检[13]。雷达类传感器在全天候工作能力方面具有天然优势[16]。然而，单一雷达模态同样难以满足无人机精确定位的需求。毫米波雷达能够直接测量目标径向速度，对高速运动目标具有较强感知能力，但其角分辨率受限于天线孔径和阵列规模，对小型无人机往往只能形成稀疏且不稳定的散射点集。在复杂城市场景中，多径效应和背景杂波进一步加剧了目标定位的不确定性，导致空间位置估计易产生抖动和漂移。相较而言，激光雷达在空间分辨率和几何建模方面表现突出，能够提供较为精确的三维点云信息。然而，其性能对气象条件和目标反射特性较为敏感，在雨雾天气或远距离观测条件下，小型无人机的点云往往急剧稀疏，难以保证时间上的连续可观测性。

上述分析表明，单模态方法的性能瓶颈并非仅源于算法设计不足，而是受限于传感器物理机理本身所导致的观测不完备性。不同模态在空间分辨率、运动感知能力及环境鲁棒性等方面存在本质差异，其失效场景往往并不重合。因此，多模态感知成为提升无人机定位与轨迹预测可靠性的必然选择。

在多模态方案中，激光雷达与毫米波雷达的融合具有显著的互补优势。激光雷达能够提供高精度的三维空间几何信息，有助于稳定无人机的空间定位与轨迹形状建模；毫米波雷达则可在远距离和恶劣天气条件下稳定工作，并通过多普勒效应直接获取目标运动信息，为动态状态估计提供可靠约束。二者在空间分辨率与速度感知能力、近距离精度与远距离鲁棒性等方面形成互补，使得融合系统在复杂环境中仍能维持连续、稳定的目标观测。

为克服上述局限，本文提出一种激光雷达和毫米波雷达融合的多模态方案，用于未授权无人机轨迹识别。该方案通过自注意力（Self-Attention）模块实现特征级融合，激光雷达能够提供集合细节信息而毫米波雷达能够提供运动速度相关信息，实现信息相互补充。自注意力机制能动态捕捉多模态数据间的相关性，提升融合效率，避免传统融合方法（如简单拼接或加权平均）的特征丢失问题。这种创新设计不仅提高了轨迹识别的准确性和鲁棒性，还在复杂环境中实现了实时处理，具有显著的理论和应用价值。本文后续将详细阐述系统框架、实验验证及性能分析。

## 1 相关研究

### 1.1 传统算法

无人机轨迹识别的传统算法包括混合估计、粒子运动模型两种主要技术路径。

混合估计算法首先应用于定位和轨迹跟踪领域，主要包括卡尔曼滤波[24]和粒子滤波[25]，通过结合多个模型或算法的状态估计，常用于处理无人机轨迹中切换动态（如从匀速到转弯）或多模态噪声的场景[12]。它通常整合如卡尔曼滤波与粒子滤波的混合，或基于交互多模型的框架，以提高轨迹预测的鲁棒性。例如[11]提出了一种基于混合阿基米德与骑手优化算法，用于无人机在复杂环境中的轨迹优化，此方案在轨迹长度、平滑度、稳定性和计算效率上表现出显著优势，特别是减少了10%的轨迹长度并提升了动态响应能力。混合估计虽能处理复杂动态，但存在计算复杂度高的问题，尤其在实时应用中，模型切换可能导致延迟。并且混合估计法对噪声敏感，如果子模型不匹配实际动态，估计误差会放大[14]。

无人机粒子运动模型将无人机视为运动过程中的运动粒子，忽略无人机的滚转运动，并使用全能量方程构建无人机轨迹模块。通过处理仿真数据以及机场低空监视雷达的实际测量数据，该方法能够在杂波干扰环境中有效跟踪无人机目标并排除飞鸟目标，从而进一步证实了其有效性和实际应用价值。粒子滤波计算密集，需要大量粒子来表示分布，导致在资源有限的无人机平台上实时性差。粒子退化是常见问题，重采样可能引入偏差，尤其在高维状态空间中效率低下。与卡尔曼滤波相比，它对运动模型假设敏感，如果模型不准（如忽略风扰动），估计会失效。在多目标场景中，粒子数爆炸式增长，进一步放大计算负担，且在有限测量下可能出现过拟合噪声的现象。

### 1.2 基于人工智能的算法

随着人工智能技术的发展，现在有越来越多的基于人工智能的非授权无人机轨迹预测的研究，现有研究主要依赖单一模态传感器，如视觉系统[15],[26]、毫米波雷达或射频信号[18],[19]。例如[27]针对视觉系统中无人机短期2-D轨迹预测的准确性和鲁棒性，提出了一种新型机制和空间-时间学习聚合器模型，通过整合空间与时间特征解决现有方法无法充分挖掘短期空间变化的问题。通过引入方位角与仰角维度分离的短期轨迹相关方法，以挖掘连续轨迹点间的空间变化并减少数据需求，提升轨迹动态捕捉能力。基于视觉的方案高度依赖环境条件，受光照变化、雾霾、夜间或恶劣天气影响，容易出现运动模糊、特征丢失或小目标检测困难，导致准确率下降。在复杂背景中，视觉方法还可能将无人机与鸟类或其它小物体混淆，尤其在远距离或高空场景下，图像分辨率不足以提供可靠特征。射频信号则通过发射高频电磁波测量反射信号，实现对无人机的距离、速度和轨迹估计。[28]提出基于自动相关监视广播技术的解决方案，并引入循环神经网络（Recurrent Neural Network, RNN)增强的卡尔曼滤波算法，通过RNN捕捉UAV机动行为和噪声水平来过滤数据噪声，实现更精确的跟踪。此外，射频信号分析也被用于检测无人机与控制器间的通信。毫米波雷达虽能在全天候条件下工作，但分辨率较低，对复杂背景中的小物体识别能力有限，且受大气吸收、多径效应、噪声干扰和范围限制，无法提供丰富的视觉细节或准确的海拔估计[18],[19]。例如，在远距离检测时，毫米波信号易受环境噪声影响，导致点云数据不准确，且对动态环境的鲁棒性不足。单一模态方法在多变环境中整体鲁棒性不足，难以满足实际应用的精度、实时性和可靠性要求。

为了解决单一模态的不足，利用多模态信息预测无人机轨迹的研究也正在开展之中。通过整合多种异构传感器数据（如视觉图像、雷达信号、LiDAR点云、惯性测量（IMU）和RF射频信号）来提升检测、跟踪和预测的精度与鲁棒性[20]-[23]。其中，[20]提出了一种名为SupFusion的新型训练策略，通过融合激光雷达和相机信息的特征，显著提升检测性能。所提出的模块能在不显著增加计算复杂度的前提下，显著提升三维检测精度。[22]提出一个名为MotionTrack 的多目标跟踪器，重点解决在拥挤和严重遮挡情况下维持目标连续轨迹的难题。针对拥挤场景，设计了交互模块，从短期轨迹中学习目标间相互影响的运动模式，以精确估计每个目标的复杂运动。针对严重遮挡导致目标丢失的问题，构建了重找回模块，利用目标的历史轨迹学习可靠的长期运动模式，从而将被中断的轨迹与正确的检测结果重新关联起来。这两个模块协同工作，使跟踪器在极具挑战性的场景下也能保持优越性能。

## 2 基于多模态融合的无人机轨迹识别框架

本实验使用毫米波雷达和激光雷达的点云数据作为神经网络的输入，首先对数据进行预处理提取无人机聚类，筛选掉背景点云。再将处理后的毫米波雷达和激光雷达数据分别输入卷积神经网络提取特征。最后通过多模态双向注意力网络得到最终的无人机定位结果。

### 2.1 雷达点云数据预处理

本研究采用 Livox Avia、LiDAR-360 以及 Oculii Eagle ETH04 77 GHz 毫米波成像采集的数据进行多模态数据采集。由于不同传感器在成像原理、空间分辨率及观测视角等方面存在显著差异，其输出数据在点云数量、视场角（Field of View, FOV）以及噪声特性等方面表现出明显不同。

Livox Avia 为非重复扫描的固态激光雷达，其视场角为朝上的约 70° 锥形视场。由于观测视角主要面向空域，该传感器采集到的点云整体较为稀疏，但具有较高的空间测距精度，且背景噪声点较少，对无人机目标点云提取的干扰相对较小。LiDAR-360 为机械旋转式激光雷达，具有更大的视场角和更高的点云密度，因此在相同时间内能够获取显著多于 Livox Avia 的空间点云数据。然而，相较于视角朝上的 Livox Avia，LiDAR-360 在采集过程中包含大量背景点云，需通过额外的背景预处理与分割步骤以减少非目标点云对后续任务的影响。

Oculii Eagle ETH04 属于 77 GHz 毫米波成像雷达，基于数字波束成形与高分辨率角度估计技术，能够在较远探测距离及复杂环境条件下稳定感知目标的速度信息与反射强度特性。相较于激光雷达，毫米波雷达对光照变化、雨雾以及遮挡等复杂环境因素具有更强的鲁棒性，但其空间分辨率相对较低，点云数据较为稀疏且噪声较大，更侧重于对目标动态特性的建模。

综上所述，针对点云数量大、背景点云占比较高的 LiDAR-360 数据，本研究设计并实施了系统的点云预处理与特征提取流程，从原始点云中定向提取无人机目标对应的点云聚类，并将其作为神经网

络的输入特征，如图 3 所示。

设原始点云序列为$\{\boldsymbol{P}_t\}_{t=1}^{T}$，其中$\boldsymbol{P}_t \in R^{N_t \times 3}$表示时间戳 t 对应的三维点云帧，$N_t$为该帧点的数量。

首先，针对数据流长度较大、帧间差异显著的特点，本研究采用帧级分块的方式对点云进行聚合。每连续 K 帧被整合为一个处理单元

$$\boldsymbol{S}_i = \{\boldsymbol{P}_{t_i}, \boldsymbol{P}_{t_{i+1}}, \ldots\ldots, \boldsymbol{P}_{t_{i+K-1}}\}. \quad (1)$$

该分段策略既能保证时间连续性，又能减少单帧噪声干扰。为了剔除异常或空值点，提高数据质量，对每帧点云执行非零掩膜筛选。

随后，为了寻找无人机目标，在一个处理单元中 $\boldsymbol{S}_i$ 对每一帧数据采用带噪声的基于密度的分层空间聚类（Hierarchical Density-Based Spatial Clustering of Applications with Noise, HDBSCAN）进行聚类划分。之所以采用 HDBSCAN 而非传统的 DBSCAN 或欧式聚类方法，主要考虑到本研究中点云数据具有明显的密度不均和噪声复杂的特点。HDBSCAN 聚类算法以最小簇大小参数$m_c$和距离阈值参数 ε 为核心超参数，将符合聚类条件的点标注为同类，输出类标签集合$\boldsymbol{L} = \{l_1, l_2, \ldots, l_n\}$，$l_i$为每个点云的类编号，其中$l_i = -1$表示噪声点，其余为目标簇编号，编号相同的点属于同一簇。

随后，为每个聚类簇提取时序特征。设时间 t 时的第 k 簇点云集合为$\boldsymbol{C}_k^t$，则计算$\boldsymbol{C}_k^t$的时间特征统计量为：

$$\mu_t = mean(\boldsymbol{C}_k^t), \quad (2)$$

$$\sigma_t = std(\boldsymbol{C}_k^t), \quad (3)$$

$$r_t = max(\boldsymbol{C}_k^t) - min(\boldsymbol{C}_k^t). \quad (4)$$

将三者拼接到一起得时间 t 时第 k 簇点云的统计特征$\boldsymbol{f}_t = [\mu_t, \sigma_t, r_t] \in R^{1\times 9}$，使用每一帧的特征构成一个处理单元 $S_i$ 中所有簇的时序特征矩阵$\boldsymbol{F}_i = [\boldsymbol{f}_{t_1}, \boldsymbol{f}_{t_2}, \ldots, \boldsymbol{f}_{t_{|\tau|}}]$。

最后，利用长短期记忆网络（Long Short-Term Memory, LSTM）对特征序列$F_i$进行时序建模与分类判别。LSTM 模型参数包括输入维度、隐藏层维度、层数与输出类别数。模型输出预测结果为 0 或者 1，表示识别为无人机目标。

通过上述处理流程，LiDAR-360 数据中的背景点与静态噪声被有效抑制，动态目标簇得以提取。该方法兼顾时间连续性与空间结构特征，为后续多模态融合网络的特征学习奠定了高质量数据基础。最终的 lidar 数据是将 Livox Avia 点云和预处理之后的 LiDAR-360 点云直接拼接。

### 2.2 基于注意力机制的多模态定位网络

对数据进行预处理，去除背景点云，得到目标无人机的点云数据之后，将其输入神经网络中进行无人机轨迹定位。本研究设计了一种用于多模态无人机轨迹预测的深度融合网络（Multi-Modal Deep Fusion Framework），整体结构由两种模态的特征提取网络与双向交叉注意力融合模块两部分组成，旨在充分利用 LiDAR 与 Radar 点云在空间几何结构与动态反射特性上的互补信息。在特征提取阶段，模型分别为 LiDAR 与 Radar 设计了独立但结构相同的特征编码器，二者有不同的权重，基于改进的 PointNet 框架实现。相比于其他网络，如 CNN、Transfomer 等 PointNet 更适合初步处理点云数据。

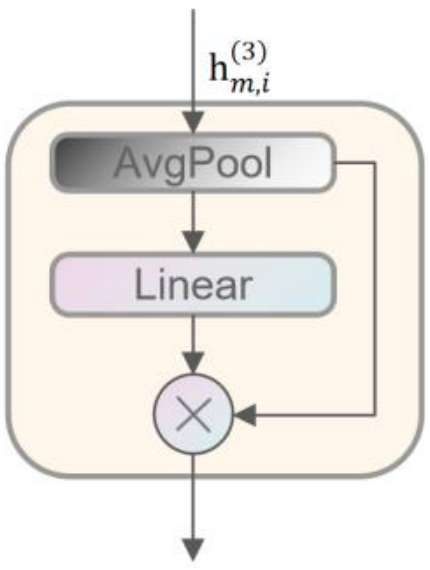

图 2　通道注意力机制

Fig.2 Block diagram of channel attention

输入点云为$\boldsymbol{X}_m = \{x_{m,1}, x_{m,2}, \ldots, x_{m,N}\} \in R^{N\times 3}$，其中 m∈{L,R}分别代表 lidar 和 radar 数据，N 为单帧数据中点的数量。模型首先通过多层感知器（MLP）进行逐点映射将三维坐标映射为高维特征（3→64→128→256），计算过程可以表示为：

$$\boldsymbol{h}_{m,i}^{(1)} = ReLU\left(\boldsymbol{W}_1 \boldsymbol{x}_{m,i} + \boldsymbol{b}_1\right), \quad (5)$$

$$\boldsymbol{h}_{m,i}^{(2)} = ReLU(\boldsymbol{W}_2 \boldsymbol{h}_{m,i}^{(1)} + \boldsymbol{b}_2), \quad (6)$$

$$\boldsymbol{h}_{m,i}^{(3)} = \boldsymbol{W}_3 \boldsymbol{h}_{m,i}^{(2)} + \boldsymbol{b}_3, \quad (7)$$

其中$\boldsymbol{W}_j$，$\boldsymbol{b}_j$分别表示可学习的权重与偏置参数，ReLU 为激活函数。

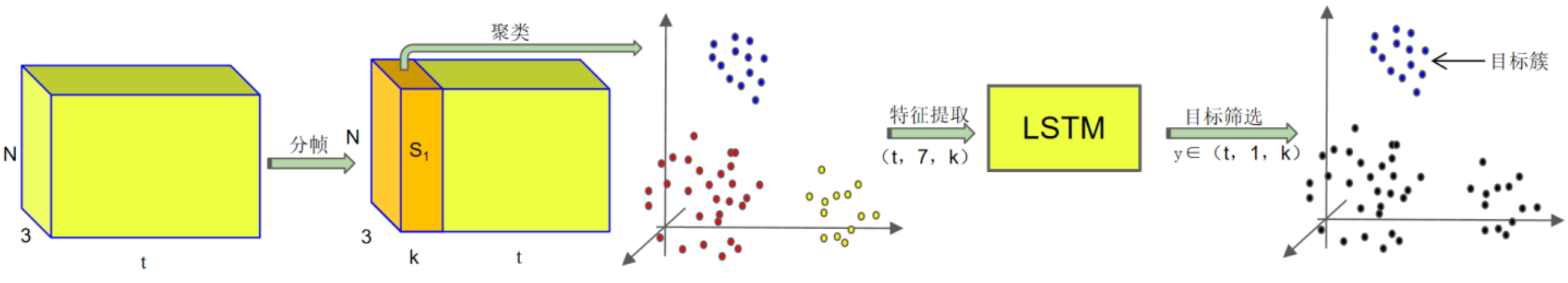

图 1　雷达数据预处理流程图

Fig.1 Illustration of pre-processing Radar data

为提升特征的表达能力与模态间一致性，PointNet Backbone 中进一步嵌入了通道注意力机制（Channel Attention），如图 2 所示。

它利用自适应全局池化提取全局通道统计信息$\boldsymbol{z} = [\boldsymbol{z}_{avg}, \boldsymbol{z}_{max}]$，随后通过两层前馈网络实现压缩与扩张映射，计算得到通道权重向量

$$\boldsymbol{a} = Sigmoid\left(\boldsymbol{W}_5 ReLU\left(\boldsymbol{W}_4 \boldsymbol{h}_{m,i}^{(3)}\right)\right), \quad (8)$$

该机制通过对特征通道进行自适应加权，使网络在不同场景下自动关注具有更强判别力的几何或语义维度。

经过注意力增强后的特征将通过全局最大池化（Global Max Pooling）操作进行汇聚，从而获得每个模态的全局特征向量$f_m \in R^{256}$。最终得到两种模态的特征表示$\boldsymbol{f}_{lidar}$和$\boldsymbol{f}_{radar}$。

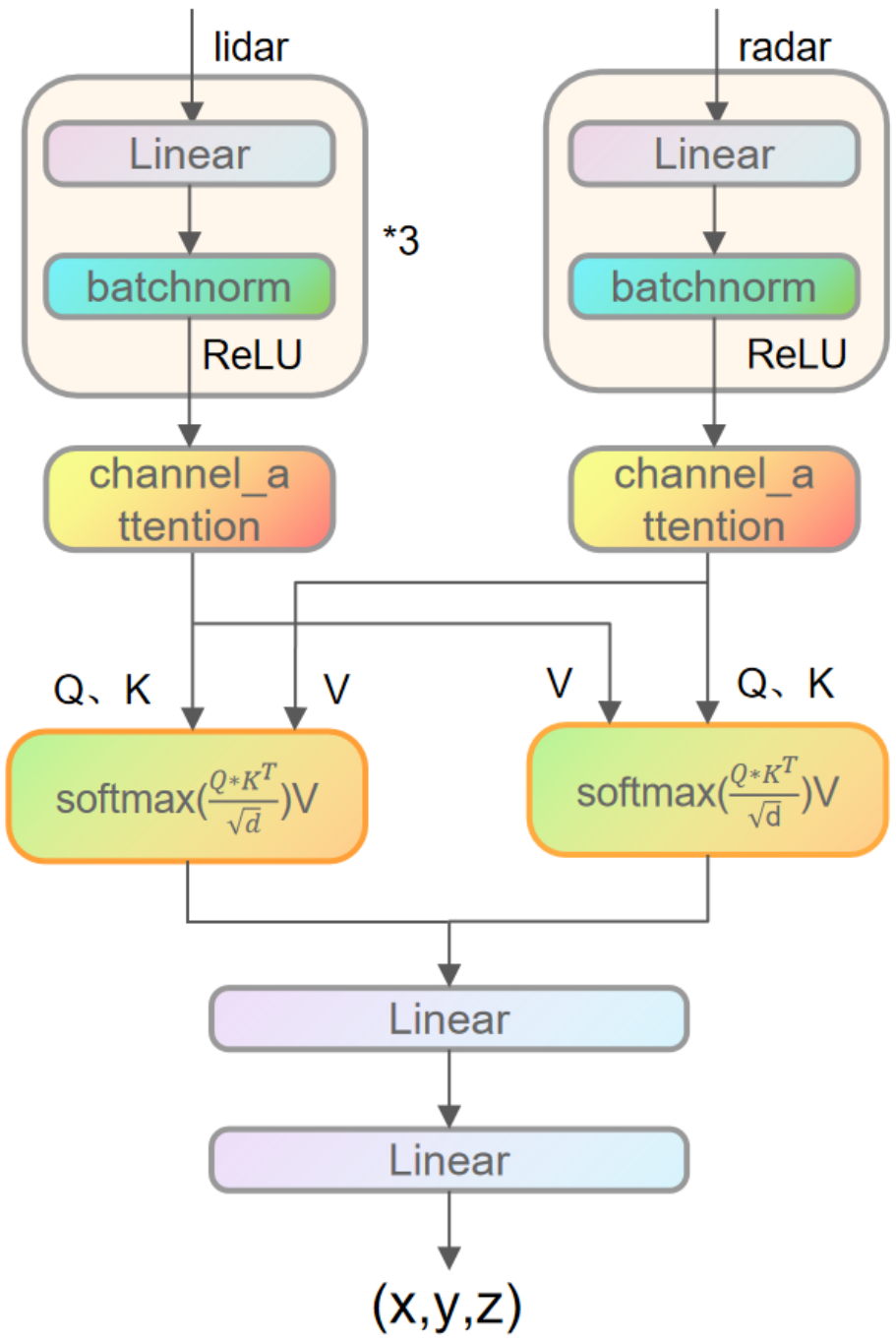

图 3　多模态融合网络结构图

Fig.3 Block diagram of feature fusion framework

提取特征之后，模型进入双向交叉注意力机制（Bidirectional Cross-Attention Mechanism）阶段，以实现两种模态间的信息互补与语义对齐。具体而言，对于 LiDAR→Radar 的交叉注意力模块，网络将 LiDAR 特征映射为查询向量$\boldsymbol{Q}_L = \boldsymbol{W}_Q \boldsymbol{f}_{lidar}$，而 radar 特征映射为键和值向量$\boldsymbol{K}_R = \boldsymbol{W}_K \boldsymbol{f}_{radar}, \boldsymbol{V}_R = \boldsymbol{W}_V \boldsymbol{f}_{radar}$。注意力的输出通过以下公式计算。

$$\boldsymbol{A}_{L\to R} = softmax\left(\frac{\boldsymbol{Q}_L \boldsymbol{K}_R^{\mathrm{T}}}{\sqrt{d}}\right)\boldsymbol{V}_R, \quad (9)$$

其中 d 为特征维度。该操作使得 LiDAR 特征能够根据 Radar 提供的语义信息进行自适应更新，从而感知另一模态的特征分布。反向方向（Radar→LiDAR）的计算过程与之对称。通过双向设计，模型可同时实现 LiDAR 与 Radar 的特征增强与信息补偿，避免单模态在融合过程中占据主导地位。

融合阶段将四个特征向量（原始 LiDAR、原始 Radar、两次交叉增强结果）进行逐元素相加：

$$\boldsymbol{f}_{fusion} = \boldsymbol{f}_{lidar} + \boldsymbol{f}_{radar} + \boldsymbol{A}_{L\to R} + \boldsymbol{A}_{R\to L}. \quad (10)$$

并输入至后续的预测网络。预测部分采用两层全连接结构（含 ReLU 激活与 Dropout 正则化），输出目标的三维坐标或位置估计结果：

$$\hat{y} = \boldsymbol{W}_p ReLU(\boldsymbol{W}_h \boldsymbol{f}_{fusiom} + \boldsymbol{b}_h) + \boldsymbol{b}_p. \quad (11)$$

通过这种融合方式，网络能够在保持结构简洁的同时，实现多模态特征的高效互补与重构。

## 2.3 神经网络输出无人机轨迹的后处理

由于神经网络的输入是单帧数据，不包含时间维度，多模态融合模型在推理过程中仍可能受到局部噪声、传感器同步误差等影响，直接输出的轨迹往往存在个别位置突变或抖动点，若不加处理，会影响整体轨迹的稳定性与速度估计的准确性。为提高预测轨迹的连续性与物理合理性，本研究在模型输出后引入了轨迹后处理步骤，包括离群点修正与时序平滑两部分。

首先，离群点修正用于检测并纠正轨迹中的异常跳变。当相邻帧间的位置差异超过预设阈值时，认为该帧点为异常点，并以其前后若干帧的平均值替代，从而在保持总体趋势不变的情况下，消除突发性的位移异常。这一步骤能够有效缓解由环境遮挡或雷达噪声导致的瞬时漂移问题，使轨迹在空间上更加连贯。

随后，采用滑动窗口方式的时序平滑进一步优化轨迹曲线。该方法通过对连续若干帧位置进行加权平均，削弱高频抖动，提升轨迹的时域连续性和平滑度。经平滑处理后的轨迹更加贴合无人机的真实运动特性，不仅在视觉上更加稳定，同时也为后续位置与速度误差的计算提供了更可靠的基础。

## 2.4 神经网络训练的损失函数设置

在本研究中，模型采用 Smooth L1 损失函数作为主要的回归优化目标。该函数在回归任务中兼具 L1 和 L2 损失的优点，定义如下：

$$L(x,y) = \begin{cases} 0.5(x-y)^2/\beta, & if\,|x-y| < \beta \\ |x-y| - 0.5\beta, & otherwise \end{cases}, \quad (12)$$

其中，$\beta$为平滑系数（通常取 1），可控制 L1 与 L2 之间的平滑过渡。当预测误差较小时，损失函数表现为二次函数形式，类似 L2 损失函数，能够平滑收敛；当误差较大时，则退化为 L1 损失函数，从而减小异常值对模型训练的不良影响。

与常规的 RMSE 相比，Smooth L1 在无人机轨

迹回归任务中具有更强的鲁棒性和数值稳定性。由于无人机轨迹数据中常包含偶发的定位跳变或噪声点，RMSE 会放大这些异常误差，从而导致训练过程不稳定；而 Smooth L1 通过在大误差区域采用线性增长的方式，有效降低了异常点的影响，使模型能更准确地学习轨迹的整体趋势与平滑变化特性。

## 3 实验与分析

### 3.1 数据介绍

本文使用的数据集为多模态专用反无人机公开数据集 MMAUD，包含立体视觉、激光雷达、毫米波雷达、音频阵列四种模态，本文选用激光雷达和毫米波雷达。数据的收集是时间连续的，在无人机飞行过程中，不同模态以不同采样频率进行数据收集，该数据集的收集超过了 1700 秒的多模态数据，每个分集包含不同型号的无人机。本文实验部分以分集 Mavic2、Mavic3、Phame 为训练集，M300 为测试集，进行相关实验。

数据集的采样频率高，时间精确到纳秒级。在具体实验过程中，激光雷达、毫米波雷达以及真值的时间并不能实现完全对齐，本文通过寻找最邻近时间的数据来实现模态之间的时间对齐。

此外，由于数据集所提供的为雷达点云数据，因此不同时间点雷达所检测到的点数也不相同，本实验对数据进行了补零操作，以保证数据输入神经网络的形状一致性。

### 3.2 神经网络训练

在本节中，我们将提出的多模态双向注意力网络与其他基线作对比。此外，还对不同损失函数设置进行对比，并对后处理进行消融实验，以验证模型有效性。神经网络训练的相关参数设置见表 1。

表 1 神经网络训练参数设置

Tab. 1 Parameter settings for neural network training

| 参数 | 数值 |
|---|---|
| 预处理分帧 K | 20 |
| 批次大小 | 32 |
| 训练周期 | 50 |
| 优化器 | Adam |

基线方法使用单模态的激光雷达数据，基于卡尔曼滤波对无人机进行时序点追踪，该方法的轨迹更加平滑，速度误差小，但是一旦出现错误可能会出现连锁反应。使用神经网络的方法在速度预测的连续性上比卡尔曼滤波差，但是定位误差有很大的降低（表 2）。

为分析损失函数对模型性能的影响，本文在相同的网络结构与训练设置下，分别采用传统的 RMSE 与 Smooth L1 Loss 进行训练，并在相同测试集上进行对比。

实验结果如表 2 所示，使用 RMSE 为损失函数的模型在测试集上得到的位置误差为 3.2063 m；相比之下，采用 Smooth L1 作为损失函数后，模型的性能显著提升，位置 RMSE 降低至 1.7858m。

表 2 无人机轨迹预测对比进行对比

Tab. 2 Performance comparison

| | 定位误差（m） | 速度误差(m/s) |
|---|---|---|
| 基线模型[23] | 2.79 | 1.73 |
| 多模态双向注意力 | 1.67 | 1.38 |

Smooth L1 损失函数在小误差区域与 RMSE 表现相似，而在大误差区域具有平滑过渡特性，不易受到异常点的影响，从而使训练过程更加稳定。该结果说明：在轨迹回归任务中，使用鲁棒性更强的 Smooth L1 损失函数能有效提升模型对异常样本的容忍度，并显著改善定位与速度估计精度。

为进一步评估预测结果后处理对模型性能的影响，本文设计了不同后处理策略的消融实验，包括无后处理、滑动平均平滑以及坏点检测结合平滑三种方式（表 3）。

在不进行任何后处理的情况下，模型输出的结果为基准，位置 RMSE 为 1.78m，速度 RMSE 为 5.26m/s。由于存在震荡的误差点，所以速度误差极大。当在测试阶段引入预测值窗口宽度为 5 的滑动平均平滑后，误差进一步降低，位置 RMSE 为 1.75m，速度 RMSE 明显下降至 1.59m/s，说明平滑操作有效抑制了预测序列中的局部震荡。此外，本文还引入了坏点检测机制（阈值设为 2m），在检测异常点后再进行平滑处理。实验结果表明，单独使用坏点检测时，位置 RMSE 为 1.61m，速度 RMSE 为 3.13m/s，位置误差最小，但是速度误差极大；而当结合坏点检测与滑动平均两种方法时，模型取得最佳性能，位置 RMSE 达到 1.67m，速度 RMSE 进一步降低至 1.38m/s。虽然相比于坏点检测位置误差略有增高，但是速度误差显著降低。

表 3 不同训练方式误差对比

Tab. 2 Comparison of Errors in Different Training Methods

| 训练方式 | 定位误差（m） | 速度误差(m/s) |
|---|---|---|
| 损失函数 RMSE（无后处理） | 3.20 | 7.36 |
| SmoothL1 损失函数（无后处理） | 1.78 | 5.26 |
| 平滑操作（窗口长度 5） | 1.75 | 1.59 |
| 坏点检测（阈值为 2） | 1.61 | 3.13 |
| 坏点检测+平滑 | 1.67 | 1.38 |

该结果表明，适当的后处理策略能够有效平滑轨迹预测结果，降低由模型震荡或噪声导致的误差，从而进一步提升整体定位与速度估计精度。

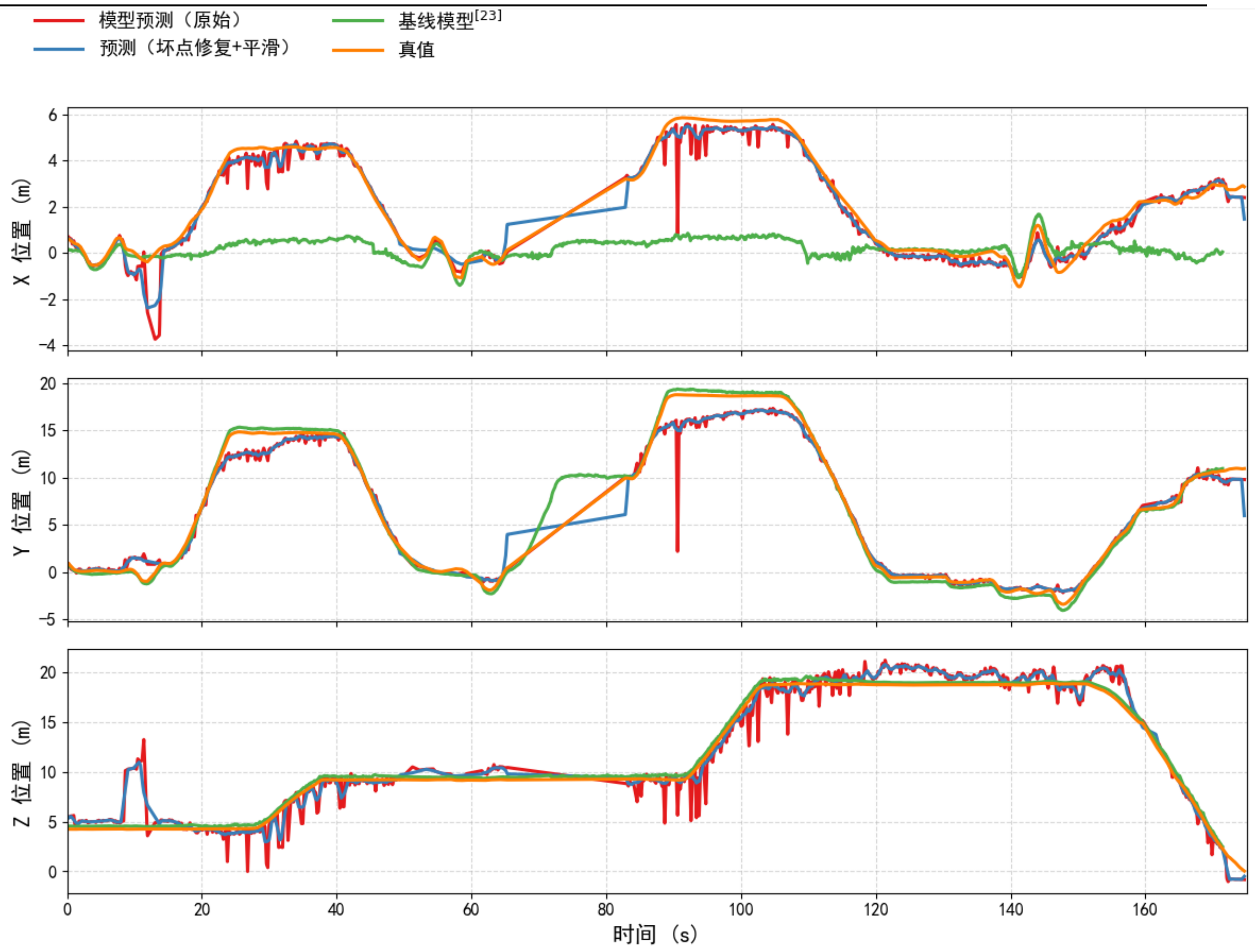

图 4　不同轨迹预测算法的结果示意图。

Fig.4　Illustration of UAV trajectory prediction results using different models.

## 4 结　论

本文提出一种基于多模态融合的非授权无人机轨迹预测的方法。不同传感器在空间、精度与鲁棒性上的互补性可通过注意力机制在特征层得到自适应整合，从而提升无人机轨迹预测的精度与稳定性。研究结果表明，使用多模态信息进行无人机定位相比于传统方法以及基于神经网络的单模态方法有明显的性能提升。这揭示了多模态感知系统中特征交互与加权选择的普遍规律，即在多源信息中动态提取可靠特征比简单拼接或固定融合更具优势。如果将来能够加入更多传感器（摄像头、麦克风阵列等），该结构可以自然的进行拓展，使用注意力机制实现模态之间的交互。

同时，本研究内容存在一些可以改进的地方：目前该模型实现的是单帧的特征融合，可以后续进行时序建模，引入 Transfomer 或 LSTM 层[29],[30]，处理连续帧信息，实现运动扩展、轨迹平滑等工作；当前方法仍依赖于较高质量的标注数据，对于极端遮挡或雷达干扰下的表现有待进一步验证。未来研究可考虑引入时空图网络或自监督学习策略[31],[32]，探索更高维度、多模态、多场景下的泛化能力，以实现更广泛应用和系统鲁棒性提升。

总之，本文的研究内容纵然存在不足，但是它在验证了基于注意力机制的多模态融合在非授权无人机定位方面的优势，为后续研究提供了参考。

## 参考文献